\pdfoutput=1

\documentclass[11pt]{article}

\usepackage[final]{acl}

\usepackage{times}
\usepackage{latexsym}

\usepackage[T1]{fontenc}

\usepackage[utf8]{inputenc}

\usepackage{microtype}
\usepackage{multicol}
\usepackage{booktabs}
\usepackage{adjustbox}
\usepackage{times}
\usepackage{amsmath}
\usepackage{pifont}
\usepackage{enumitem}
\usepackage{latexsym}
\usepackage{lipsum}
\usepackage{graphicx}
\usepackage[T1]{fontenc}
\usepackage[toc,page]{appendix}

\usepackage[utf8]{inputenc}

\usepackage{microtype}

\usepackage{subfig}
\newcommand{\cmark}{\ding{51}}%
\newcommand{\xmark}{\ding{55}}%

\NewDocumentCommand{\lifu}{ mO{} }{\textcolor{OrangeRed}{\textsuperscript{\textit{Lifu}}\textsf{\textbf{\small[#1]}}}}
\NewDocumentCommand{\lichao}{ mO{} }{\textcolor{blue}{\textsuperscript{\textit{Lichao}}\textsf{\textbf{\small[#1]}}}}
\title{Membership Inference Attacks on Knowledge Graphs}

\author{Yu Wang$^1$, \textbf{Lifu Huang}$^2$, \textbf{Philip S. Yu}$^1$, \textbf{Lichao Sun}$^3$ \\
        1 University of Illinois at Chicago, 2 Virginia Tech, 3 Lehigh University\\
        \texttt{ywang617,psyu@uic.edu} \\ \texttt{warrior.fu@gmail.com}\\\texttt{james.lichao.sun@gmail.com}}

\begin{document}
\maketitle
\begin{abstract}
Membership inference attacks (MIAs) infer whether a specific data record is used for target model training. MIAs have provoked many discussions in the information security community since they give rise to severe data privacy issues, especially for private and sensitive datasets. Knowledge Graphs (KGs), which describe domain-specific subjects and relationships among them, are valuable and sensitive, such as medical KGs constructed from electronic health records. However, the privacy threat to knowledge graphs is critical but rarely explored.
In this paper, we conduct the first empirical evaluation of privacy threats to knowledge graphs triggered by knowledge graph embedding methods (KGEs). We propose three types of membership inference attacks: transfer attacks (TAs), prediction loss-based attacks (PLAs), and prediction correctness-based attacks (PCAs), according to attack difficulty levels. 
In the experiments, we conduct three inference attacks against four standard KGE methods over three benchmark datasets. In addition, we also propose the attacks against medical KG and financial KG. The results demonstrate that the proposed attack methods can easily explore the privacy leakage of knowledge graphs. 

\end{abstract}

\section{Introduction}
Knowledge graphs (KGs) are critical data structures to represent human knowledge, and serve as resources for various real-world applications, such as recommendation \cite{zhang2016collaborative, gong2021smr}, question answering \cite{yih2015QA, liu2018t, hasan2016clinical}, disease diagnosis~\cite{chai2020diagnosis, fang2019diagnosis}, etc. Typical KGs are the collections of triples \textit{(head entity, relation, tail entity)} that represent the domain knowledge. For example, the triple \textit{(Heart Attack, subClass\_of, Heart Diseases)} states the fact that heart attack is one kind of heart disease.
Knowledge graph embedding methods (KGEs)~\cite{bordes2013transE,wang2014transH,yang2014distmult, trouillon2016complex} have been explored and developed to learn low dimensional representations from knowledge graphs. However, KGs constructed from social platforms, financial or medical institutions usually contain sensitive information. For example,~\citet{gong2021smr} apply KGEs on a KG constructed from electronic medical records. A severe privacy leakage will incur if the adversary can collect such information of private KGs through inference attacks \cite{qian2016privacy}.

\begin{table}[t]
    \caption{The ability of adversary's knowledge for different attacks}
    \label{tab:know_ana}
\begin{adjustbox}{width=2.5in,center}
    \centering
    \begin{tabular}{cccc}
    \hline
    & TAs & PLAs & PCAs \\
\hline    
    Shadow Dataset: & \cmark & \xmark & \xmark\\ 
    Plausibility Score: & \cmark & \cmark & \xmark  \\
    Prediction Label: & \xmark& \xmark & \cmark \\
    \hline
    \end{tabular}
    \end{adjustbox}
\vspace{-15pt}
\end{table}

\begin{table*}[t]
    \caption{The basics information of KGE models}
    \label{tab:KGEs}
    \begin{adjustbox}{width=4.5in,center}
    \centering
    
    \begin{tabular}{cccc}
    \hline
         Model& Scoring Function &  Loss
         Function  \\
         \hline
         TransE& $-||\mathbf{h+r-t}||_{L_1/L_2}$ & margin-based loss\\
         TransH& $-||\mathbf{(h-w_r^Thw_r)+r-(t-w_r^Ttw_r)}||_{L_1/L_2}$ & margin-based loss\\
         DistMult & $<\mathbf{h,r,t}>$ & logistic loss\\
         ComplEx & $\mathcal{R}(\mathbf{<h, r, \Bar{t}>})$ & logistic loss\\
         \hline
    \end{tabular}
    \end{adjustbox}

\end{table*}

A series of recent studies~\cite{fredrikson2015modelinversion,goodfellow2014explaining,sun2020natural,he2021model} has shown that machine learning~(ML) models are vulnerable to various security and privacy attacks.
One specific attack is the membership inference attack (MIA) that can infer whether a particular data record was in the training dataset of a given ML model, thereby causing data breach issues of sensitive user information. The explorations of various MIAs have been proposed in recent literature
\cite{hu2021membership,hu2021source,long2017towards,shokri2017membership,salem2018ml,yeom2018privacy,song2019privacy,li2020membership}.

Nevertheless, most prior studies in this direction focused on image datasets. \cite{duddu2020quantifying} examines the privacy threat when node embeddings are released. ~\cite{olatunji2021membership,he2021node} study the privacy issues concerning confidence vectors from graph neural networks (GNNs). These attacks also implicitly assume access to node embeddings since the confidence vectors result from applying softmax function on node representations from the final layer of GNNs. Node embeddings can be easily retrieved by solving a system of linear equations.

Unlike MIAs to heterogeneous and homogeneous graphs, practical attacks to KGE models become more challenging because of 1) \textit{realistic black-box setting} of KGE attacks and 2) \textit{free-scale predictions}. In real-world scenarios, attackers fail to access the model parameters on black-box settings~\cite{shokri2017membership,salem2018ml} including entity and relation embeddings of KGs. On the other hand, the output score of KGE models entangles the entity and relation embeddings together, which eliminates the possibility of back retrieving entity and relation embeddings from outputs. This realistic limitation required well-designed feasible solutions for KG attacks. Furthermore, unlike the confidence vector, the prediction scores of KGE models are free-scale and uncalibrated~\cite{tabacof2019probability}. The prediction score distribution of different KGE models may shift significantly. Such characteristics highly increase the complexity of attacks.

To address these challenges, we propose three types of MIAs based on the scope of the adversary's knowledge (as shown in Table~\ref{tab:know_ana}). transfer attacks(TAs) train the adversary directly on output scores with the assistance of the shadow dataset, which has similar distribution with the target dataset. Prediction loss-based attacks(PLAs) measures the output scores by measuring the loss of member and non-member samples, and prediction correctness-based attacks(PCAs) make predictions only by querying target for output labels. 
Our contributions can be summarized as follows: 1) To our best knowledge, this is the first empirical study to investigate the membership leakage in KGE models; 2) We systematically define the threat model of MIAs against KGE models; 3) We propose three different attack techniques based on the scope of the adversary's knowledge, and perform the proposed MIAs in the most challenging and realistic black-box setting; 4) We perform MIAs against four standard KGE models on three benchmark KG datasets. We also conduct attacks against sensitive medical and financial KGs to demonstrate that this problem is realistic and critical. Our extensive experimental results reveal that KGEs are vulnerable to MIAs and call for future privacy-preserving knowledge graph embedding methods.

\section{Knowledge Graph}
This section introduces the formal definition of KG and general architecture of KGE.

\noindent\textbf{Definition of KG.}
KG is a directed heterogeneous graph $\mathcal{K}=\{\mathcal{E}, \mathcal{R}, \mathcal{T}$\}, where $\mathcal{E}$ and $\mathcal{R}$ denote the set of entities and relation types respectively. $\mathcal{T}=\{\tau: (h,r,t)\in \mathcal{T}|h,r\in \mathcal{E}, r\in \mathcal{R}\}$ is the set of fact triples.

\noindent\textbf{Knowledge Graph Embedding.}
Knowledge graph embedding (KGE) methods aim to map symbolic entities and relations into continuous embedding space \cite{bordes2013transE,wang2014transH,yang2014distmult, trouillon2016complex}. 
To be specific, KGEs define a score function $S(\tau)$ to measure the plausibility for each triple $\tau$. 
We summarize the different score and loss functions of KGE models in Table~\ref{tab:KGEs}.
Some KGEs, e.g., TransE \cite{bordes2013transE} and TransH \cite{wang2014transH}, apply margin-based loss to learn the entities' and relations' embeddings:
\vspace{-8pt}
\begin{align*}
        \mathcal{L} = \sum_{(\tau)\in \mathcal{T}}\sum_{(\tau')\in \mathcal{T'}}max(\gamma + S(\tau)-S(\tau'),0),
        \vspace{-10pt}
\end{align*}
where $\mathcal{T'}$ is a set of false facts with corrupted entities or relations.
Some KGEs, e.g., Distmult \cite{yang2014distmult} and ComplEx \cite{trouillon2016complex}, use logistic-based learning objective to ensure that true facts have lower scores than false facts:
\vspace{-8pt}
\begin{align*}
\begin{split}
       \mathcal{L} = &\sum_{\tau\in \mathcal{T},\tau' \in \mathcal{T}'} log(1+e^{(S(\tau)})+ log(1+ e^{(-S(\tau')}). \vspace{-10pt}
\end{split}
\end{align*}

\section{Membership Inference Attacks on Knowledge Graph}
In this section, we first define the problem of membership inference attacks on the knowledge graph. Then, we discuss a threat model regarding the adversary's background knowledge. Last, based on the scope of the adversary's knowledge, we propose three privacy attack methodologies.

\subsection{Problem Definition}
The objective of an adversary is to determine whether a triple $\tau$ of a knowledge graph $\mathcal{K}$ is used to train a KGE model. Formally, given a candidate triple $\tau \in \mathcal{K}$, the target model $\mathcal{M}_{T}$, and the background knowledge $\Omega$ of the adversary, the membership inference attack $\mathcal{A}$ can be defined as follows:
    \vspace{-10pt}
\begin{equation}
    \mathcal{A}: \tau, \mathcal{M}_{T}, \Omega \; \rightarrow \{0, 1\},
    \vspace{-10pt}
\end{equation}
where $1$ represents that the candidate triple $\tau$ is in $\mathcal{M}_{T}$'s training set and $0$ otherwise.

\subsection{Threat Model}

In this paper, we assume that the adversary only has black-box access to the target KGE model $\mathcal{M}_T$, which is trained on the private knowledge graph $D_T$. In this case, the adversary can only obtain the outputs of the KGE model $\mathcal{M}_T(\tau)$, where $\tau$ is any candidate triple in the knowledge graph $\mathcal{K}$. This setting is not only challenging but very realistic for membership inference attacks \cite{salem2020updates, salem2018ml, shokri2017membership}. In this circumstance, we first characterize the adversary's knowledge along three dimensions in Table \ref{tab:know_ana} and then propose three different MIAs based on the scope of the adversary's knowledge.

\begin{figure}[t]
    \centering
    \includegraphics[height=2in, width=\columnwidth]{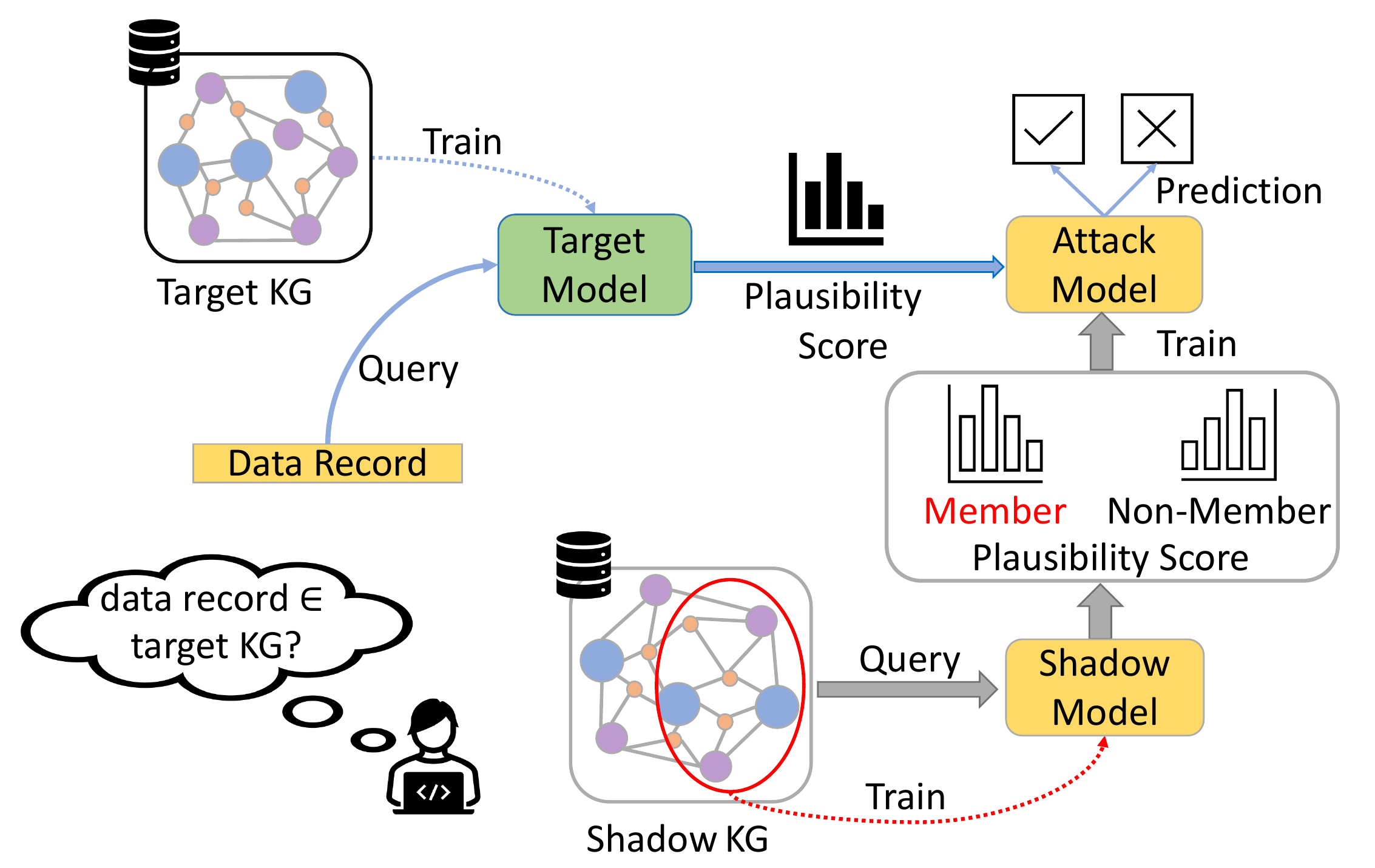}
    \caption{Overview of the transfer attacks on KGs}
    \label{fig:attack1}
\vspace{-15pt}
\end{figure}

\noindent\textbf{Shadow dataset} refers to a dataset $D_{S}$ whose distribution is similar to the target dataset $D_{T}$. There are many scenarios where an adversary can access a similar dataset. For example, in federated learning paradigm~\citet{sui2020feded}, each clinical institute has a subset of medical records. Each exchange institute stores a set of stock trading records in financial applications. Many inference attack methods \cite{shokri2017membership,salem2018ml} assume access to the shadow dataset that helps the adversary train an auxiliary shadow model $\mathcal{M}_S$, which can mimic the behaviors of the target model.


\noindent\textbf{Output knowledge} refers to the knowledge of the target model's output. The objective of the triple classification task for the KGE models is to infer whether a given triple $\tau = (h, r, t)$ is correct or corrupted. For each triple $\tau$, the KGE models will use the score function $S(\tau)$ to measure the plausibility of this triple and predict as corrupted fact if the score is above the preliminary threshold and vice versa. Thus, the output can be categorized into two different types according to the amount of information accessible to the adversary:

\begin{itemize}[noitemsep,topsep=0pt]
    \item $\textbf{Plausibility score}$ indicates the output of score function $S(\tau)$ of the target model.
    \item $\textbf{Prediction label}$ stands for the target model's inference about whether a triple is correct or corrupted. This reveals the minimum amount of information to the adversary.
\end{itemize}

Based on whether the adversary has access to the above knowledge, we propose three inference attacks: transfer attacks, prediction loss-based attacks, and prediction correctness-based attacks.

\subsection{Transfer Attacks}

We first propose the transfer attacks (TAs), assume the adversary has access to both the shadow dataset and the plausibility score. The proposed TAs build upon the standard $\textit{shadow model approach}$ \cite{shokri2017membership}. Figure~\ref{fig:attack1} shows TAs contain three main steps as follows.

\noindent\textbf{Shadow model training.}
The objective of the shadow model is to replicate the functionality of the target model. 
To achieve this goal, the adversary uses the same model architecture as $\mathcal{M}_T$ to train the model on the shadow dataset.
Furthermore, the adversary applies the same training algorithm for the shadow model. 
For more details, the adversary randomly splits the shadow dataset $D_{S}$ into two disjoint subsets: $D_{S}^{train}$ and $D_{S}^{test}$, where the $D_{S}^{train}$ is used to train the shadow model.

\noindent\textbf{Attack model training.}
The attack model's goal is to distinguish members and non-members of the training set. To train the attack model $\mathcal{A}$, the adversary first queries triples from $D_{S}^{train}$ and $D_{S}^{test}$ to the shadow model for plausibility scores, respectively. The plausibility scores of triples in $D_{S}^{train}$ are labeled as $1$, indicating memberships of the training set and $0$ otherwise. Then the adversary trains a MLP classifier that takes plausibility scores as inputs and predicts their membership labels.

\noindent\textbf{Membership inference.}
After training the attack model, the adversary uses it for membership inference attacks. To infer the membership of a given candidate triple $\tau$, the adversary first queries the target model $\mathcal{M}_{T}$ for the candidate triple's plausibility score. Then, the adversary feeds the plausibility score of the triple $\tau$ into the attack model $\mathcal{A}$ to obtain the membership status.

\subsection{Prediction Loss-based Attacks}
\begin{figure}
    \centering
    \includegraphics[height=1.2in, width=\columnwidth]{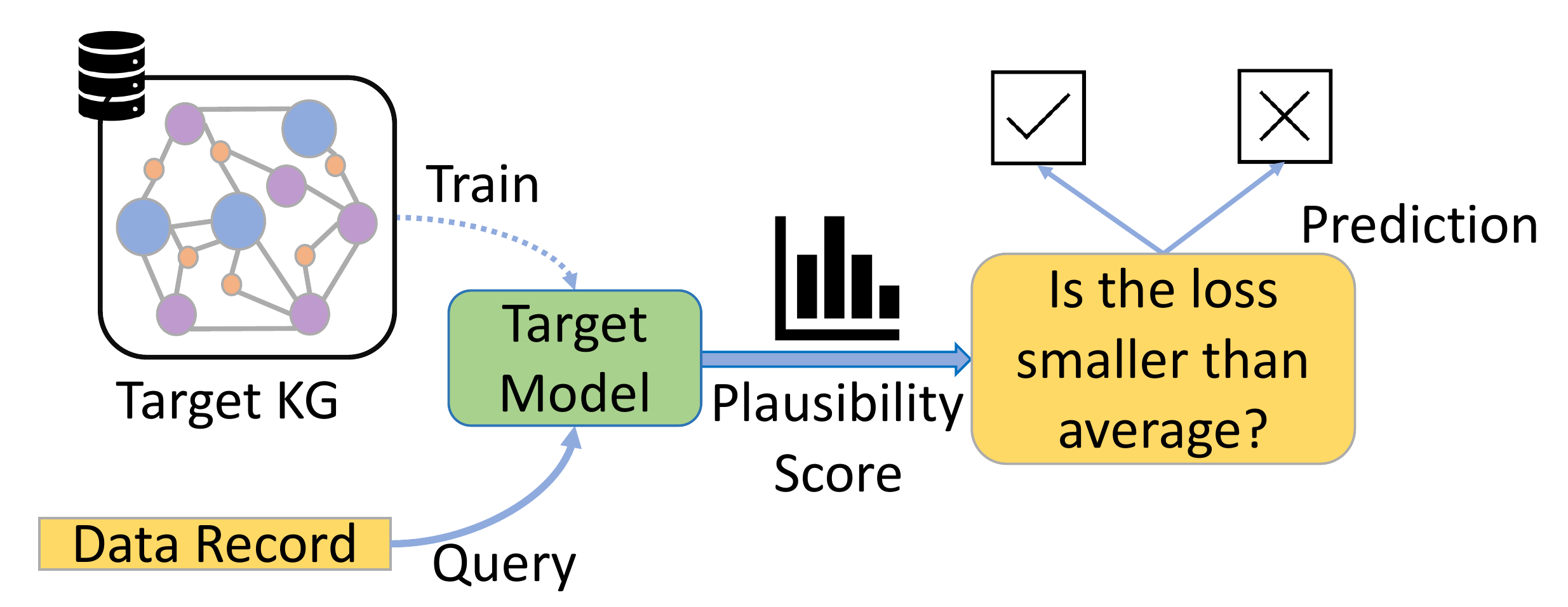}
    \caption{Overview of the prediction loss-based attacks on knowledge graphs}
    \label{fig:attack2}
\vspace{-15pt}
\end{figure}

Our second proposed attack method is prediction loss-based attack (PLA), which assumes the adversary has no access to the shadow dataset $D_S$. The adversary can only query the target model for plausibility scores in this situation. This is a more practical setting than TAs due to the less information required by the adversary. In real life, it is almost impossible or too expensive for adversaries to collect such a dataset with similar data distribution as the private training set.
To attack the model without a shadow dataset, many metric-based methods are introduced \cite{yeom2018privacy,song2019privacy,leino2020stolen,song2021systematic}. 
Inspired by the above projects, we propose the prediction loss-based attack method on the knowledge graphs. 
As shown in Figure~\ref{fig:attack2}, the adversary first queries a set of candidate triples to the target model about the plausibility scores. Second, the adversary uses a loss function as the metric to compute the loss value of each triple and then calculate the average loss over all data samples. Next, the adversary predicts a triple $\tau$ as a training dataset member if the prediction loss is below average. The PLA method can be formalized as:
\vspace{-10pt}
\begin{equation}
    \mathcal{A}(\hat{p}(\tau)) = I(\mathcal{L}(\hat{p}(\tau))\leq \delta),
    \vspace{-10pt}
\end{equation}
where $\hat{p}(\tau)$ is the plausibility score, $\delta$ is the average loss and $I(\cdot)$ is the indicator function. Compared with TAs, PLAs require less computational resources and attack knowledge.

\subsection{Prediction Correctness-based Attacks}
\begin{figure}
    \centering
    \includegraphics[height=1.2in, width=\columnwidth]{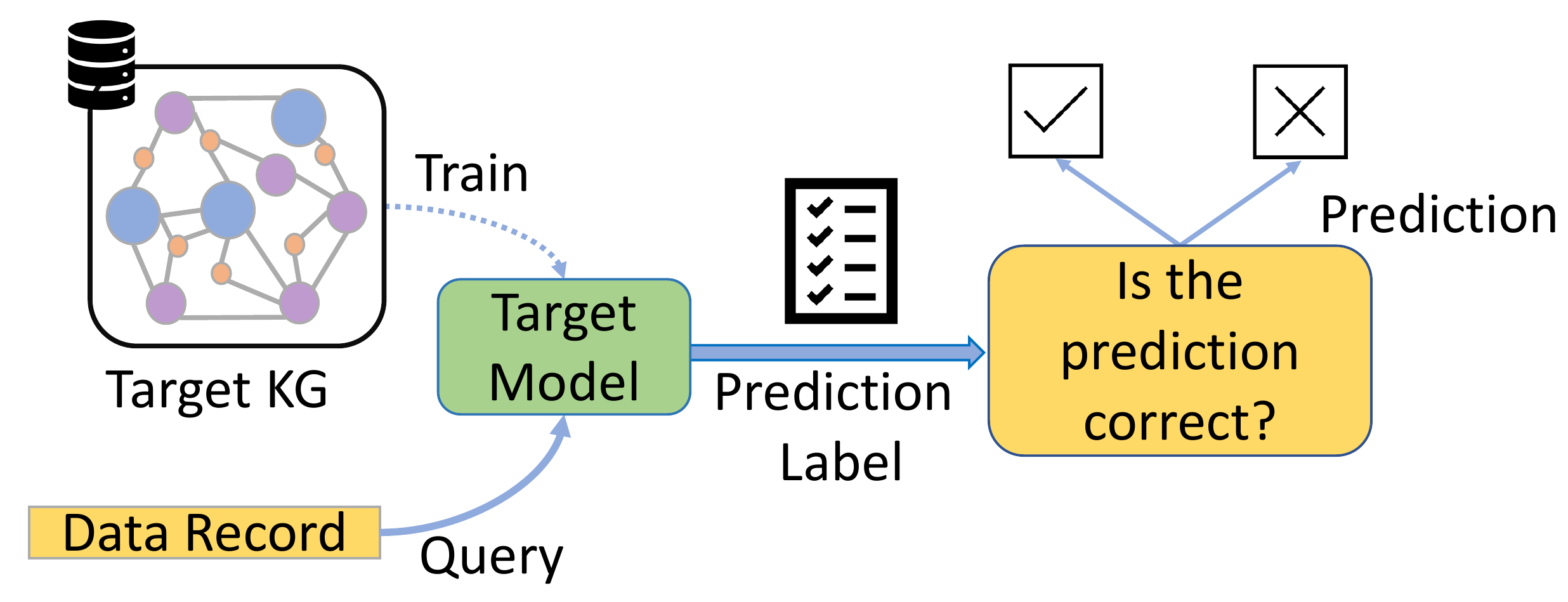}
    \caption{Overview of the prediction correctness-based attacks on knowledge graphs}
    \label{fig:attack3}
\vspace{-15pt}
\end{figure}
In contrast to TAs and PLAs, we restrict the adversary's knowledge further. The adversary only has access to the predicted hard labels for privacy attacks, namely prediction correctness-based attacks (PCAs). It is the most challenging setting due to the least amount of information of the adversary. Apparently, this setting is the most realistic among all three inference attacks. 
As shown in Figure~\ref{fig:attack3}, given a triple and its label $(\tau, y)$, the adversary queries the target model and observes whether the prediction label is correct. The adversary predicts the misclassified triples as a non-member of the target training dataset and vice versa. The attack method can be formalized as follows:
\vspace{-10pt}
\begin{equation}
    \mathcal{A}(\hat{p}(\tau); y) = I(\hat{p}(\tau) == y),
    \vspace{-10pt}
\end{equation}
where $\hat{p}(\tau)$ is the prediction label from the target model. The intuition behind is that the target model can correctly predict training data, but they may not generalize well on test data.


\begin{table*}[t]
\caption{Experimental results on TAs.}\label{tab:transfer_attack}
\begin{adjustbox}{width=\textwidth,center}
    \centering
\begin{tabular}{|c|cccc|cccc|cccc|cccc|cccc|cccc}
\hline
{}& \multicolumn{4}{|c|}{WN18RR} &
\multicolumn{4}{c|}{FB15K237} &
\multicolumn{4}{c|}{NELL-995} &
\multicolumn{4}{c|}{DDB14}&
\multicolumn{4}{c|}{FinKG}\\
\hline
Model & Acc & F1 &Precision & Recall& Acc & F1 &Precision & Recall& Acc & F1 &Precision & Recall& Acc & F1 &Precision & Recall&Acc & F1 &Precision & Recall\\ 
\hline

TransE  &    90.48 &   90.84 &          87.52 &       94.43 &      83.57 &     84.32&            80.65 &         88.33 &      76.10 &     73.92 &            81.37 &         67.71 &    95.37 &  95.31 &         96.59 &      94.06  &        99.99 &       99.99 &                  100 &           99.98 \\
TransH   &    91.37 &   91.79 &          87.53 &       96.49 &      84.37 &     85.85 &            78.40 &          94.86 &      85.84 &     85.21 &            89.17 &         81.60 &   95.76 &  95.69 &         97.13 &      94.30 &        99.96&       99.96 &                  100 &           99.92 \\
ComplEx  &    94.82 &   95.07&          90.61 &           100 &       84.04 &     86.05 &            76.41 &         98.47 &      92.02 &      92.60 &            86.30 &         99.89 &   99.72 &  99.72 &         99.45 &          100 &        93.86 &       94.22 &               89.07 &               100 \\
DistMult &    94.53 &   94.81 &          90.14 &           100 &      75.66 &     79.75 &            68.27 &         95.86&      87.70 &     89.01 &            80.45 &           99.62 &   99.12 &  99.13 &         98.28 &          100 &        87.09 &       88.56 &               79.48 &               100 \\
\hline
{Average Acuracy}& \multicolumn{4}{|c|}{$93.02\%$} &
\multicolumn{4}{c|}{$81.91\%$} &
\multicolumn{4}{c|}{ $85.42\%$} &
\multicolumn{4}{c|}{$97.49\%$}&
\multicolumn{4}{c|}{$95.22\%$}\\
\hline
\end{tabular}
\end{adjustbox}
\vspace{-15pt}
\end{table*}

\begin{figure*}
    \centering
    \includegraphics[width=\textwidth]{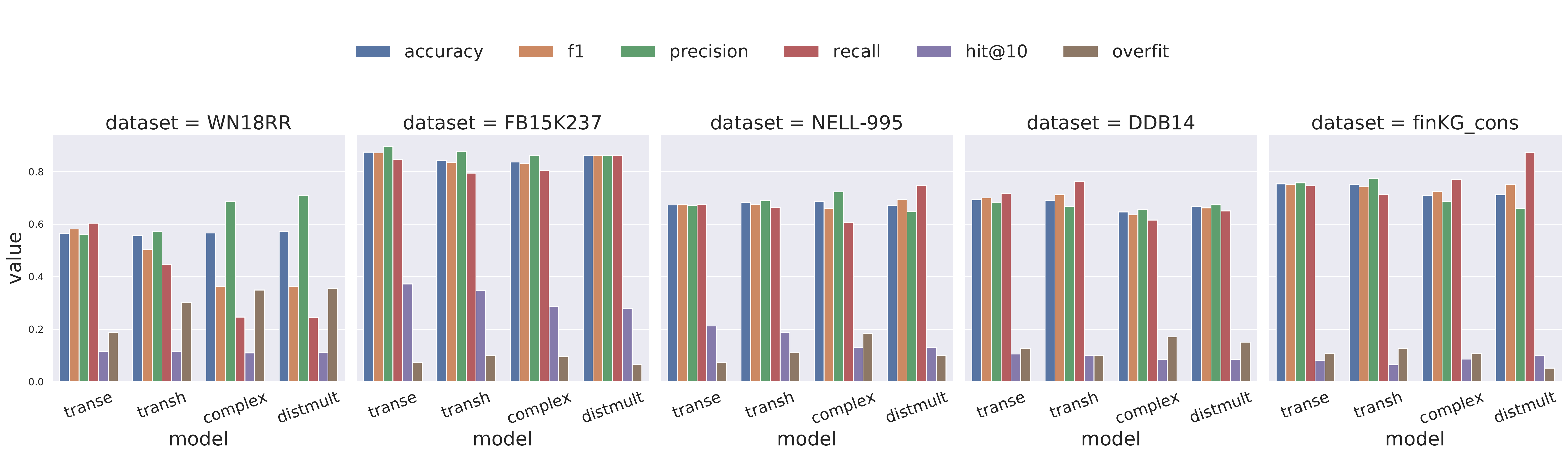}
    \caption{The performance analysis of target models on target datasets}
    \label{fig:ori_performance}
\vspace{-5pt}
\end{figure*}

\section{Evaluation}
In this section, we first introduce the experimental setup and then perform a comprehensive analysis of MIAs against the four standard KGE models on five datasets to evaluate the performance of the proposed TAs, PLAs, and PCAs.

\subsection{Experimental Setup}
\noindent\textbf{Datasets.}
To evaluate the performance of the proposed attacks, we use three benchmark datasets for the KGE tasks, including WN18RR, FB15K237, and NELL-995. In order to support our claim that the privacy leakage issue is severe and realistic, we introduce two private knowledge graphs: DDB14~\cite{wang2021relational} and FinKG.  DDB14 is a set of triples that are extracted from the Disease Database. This medical database records terminologies and concepts of diseases, symptoms, related drugs, and corresponding relationships. FinKG is a new financial knowledge graph, which extracted from Tushare\footnote{http://www.tushare.org/}. FinKG encodes the information of stocks and stockholders, such as code, concepts of stocks, name, amount of holding stocks, date of purchase for stockholders. We can train the KGE models to predict whether a user in the system would purchase a specific stock. 
Table~\ref{tab:data_ana} summarizes the statistics of these datasets.

For fair comparisons, we follow the same settings
as a prior study \cite{shokri2017membership}. Specifically, we first randomly split the knowledge graph into two equal disjoint subsets $D_{T}$ and $D_{S}$ with similar data distributions for training and testing, respectively. We further split two equal disjoint subsets $D_S^{train}$ and $D_S^{test}$ of $D_S$ with the first one being used for training and the second one being used for testing of shadow model, which is also the same process for $D_{T}$.
Note that $D_{T}$ also serves as an evaluation dataset of the attack model. Specifically, for FinKG, we extract subgraphs from Shanghai Stock Exchange as the target dataset while subgraphs from Shenzhen Stock Exchange as the shadow dataset. We denote this split as finKG\_cons in experiments. This setting is more realistic since, in many real-world applications, such as Federated Learning~\cite{sui2020feded}, a local client has access to its own subgraph, where privacy leakage will be triggered if a malicious server aims to infer the data memberships from each local client.

\noindent\textbf{Models.}
We use four different KGE methods to construct both shadow and target models, including TransE~\cite{bordes2013transE}, TransH~\cite{wang2014transH}, ComplEx~\cite{trouillon2016complex}, and DistMult~\cite{yang2014distmult}. We followed the hyperparameter configuration provided by OpenKE \cite{han2018openke} and chose the best model to attack.

\noindent\textbf{Metrics.}
Four standard attack metrics are used to evaluate the performance of the proposed attacks, including accuracy, F1 score, precision, and recall. The precision score measures the fraction of triples inferred as members that are indeed members, and the recall measures the fraction of triples in the training set that are inferred as members. 

\begin{table}[t]
    \caption{Dataset statistics}
    \label{tab:data_ana}
\begin{adjustbox}{width=2.1in,center}
    \centering
    \begin{tabular}{|c|ccc|}
    \hline
    & Relations &Entities &Triples\\
\hline    
    WN18RR & 11 & 40,943 & 92,973\\ 
    FB15K237 & 237 & 14,541 &  310,116\\
    NELL-995 & 200 & 75,492 & 154,213\\
    DDB14& 14 & 9,203 & 66,841\\
    FinKG&13&15,847&32,404\\
    \hline
    \end{tabular}
    \end{adjustbox}
\vspace{-15pt}
\end{table}

\begin{table*}[]
\caption{Experimental results on PLAs with logistic-based loss.}\label{tab:pla_softplus}
\begin{adjustbox}{width=\textwidth,center}
    \centering
\begin{tabular}{|c|cccc|cccc|cccc|cccc|cccc|cccc}
\hline
{}& \multicolumn{4}{|c|}{WN18RR} &
\multicolumn{4}{c|}{FB15K237} &
\multicolumn{4}{c|}{NELL-995} &
\multicolumn{4}{c|}{DDB14}&
\multicolumn{4}{c|}{FinKG}\\
\hline
Model & Acc & F1 &Precision & Recall& Acc & F1 &Precision & Recall& Acc & F1 &Precision & Recall& Acc & F1 &Precision & Recall&Acc & F1 &Precision & Recall\\ 
\hline

TransE   &    89.63 &   90.53 &          83.30 &       99.13 &       82.87 &     84.60 &            76.82 &         94.13&      72.45 &     74.18 &            69.79 &         79.15 &   94.76&  94.95 &         91.76&      98.36&        99.80&       99.80 &              99.60 &               100 \\
TransH   &    89.22 &   90.25 &          82.39 &       99.77 &      84.41 &     85.86&            78.52&         94.72 &      84.23 &     85.03 &            80.90 &         89.60 &    95.38 &  95.50 &         93.08 &      98.04 &        99.98 &       99.98 &              99.97 &               100 \\
ComplEx  &    93.56 &   93.95 &          88.60 &       99.98 &      72.65 &     78.18 &            65.03 &         98.02 &      84.04 &     86.22 &            75.87 &         99.85 &   94.00 &  94.34 &         89.30 &        99.99 &        86.66 &       88.22 &              78.98 &           99.91 \\
DistMult &    93.01 &   93.46 &          87.79 &       99.92 &      67.13 &     74.50 &            60.85 &         96.03 &      81.78 &     84.53 &            73.43 &         99.60 &   92.60 &  93.10 &         87.26 &      99.78 &        83.58 &       85.85 &              75.42 &           99.63\\
\hline
{Average Acuracy}& \multicolumn{4}{|c|}{$91.35\%$} &
\multicolumn{4}{c|}{$76.76\%$} &
\multicolumn{4}{c|}{ $80.62\%$} &
\multicolumn{4}{c|}{$94.18\%$}&
\multicolumn{4}{c|}{$92.50\%$}\\
\hline
\end{tabular}
\end{adjustbox}
\end{table*}

\begin{table*}[h]
\caption{Experimental results on PLAs with margin-based loss.}\label{tab:pla_margin}
\begin{adjustbox}{width=\textwidth,center}
    \centering
\begin{tabular}{|c|cccc|cccc|cccc|cccc|cccc|cccc}
\hline
{}& \multicolumn{4}{|c|}{WN18RR} &
\multicolumn{4}{c|}{FB15K237} &
\multicolumn{4}{c|}{NELL-995} &
\multicolumn{4}{c|}{DDB14}&
\multicolumn{4}{c|}{FinKG}\\
\hline
Model & Acc & F1 &Precision & Recall& Acc & F1 &Precision & Recall& Acc & F1 &Precision & Recall& Acc & F1 &Precision & Recall&Acc & F1 &Precision & Recall\\ 
\hline

TransE   &    88.50 &   89.37 &          83.10 &        96.66 &      71.11 &     76.73 &            64.23 &         95.27 &      69.80 &     70.79&            68.54 &         73.19 &   81.32 &  82.99 &         76.16 &      91.18 &         91.52 &       92.18 &              85.50 &               100 \\
TransH   &    89.96 &   90.76 &          84.07 &       98.60 &      75.67 &     79.49 &             68.70 &         94.31&      76.56 &     78.39 &            72.72&         85.02 &   81.69&  83.37 &         76.37 &      91.79 &         89.22 &       90.27 &               82.27 &               100 \\
ComplEx  &    91.86 &   92.47 &          86.0 &       99.98&      69.79 &     76.50 &            62.60 &         98.32 &      82.54 &     85.12 &            74.16 &         99.89 &   89.55 &  90.54 &         82.72 &        99.99 &        83.32 &       85.70&              75.00 &           99.95 \\
DistMult &    92.47 &   92.99 &          86.97&       99.91 &      66.04 &     73.89 &             60.01 &         96.12 &      81.72 &     84.49 &            73.34 &         99.65 &   88.87 &  89.96 &         81.91 &      99.78 &        82.42&       84.98 &              74.16 &            99.50\\
\hline
{Average Acuracy}& \multicolumn{4}{|c|}{$90.69\%$} &
\multicolumn{4}{c|}{$70.65\%$} &
\multicolumn{4}{c|}{ $77.65\%$} &
\multicolumn{4}{c|}{$85.35\%$}&
\multicolumn{4}{c|}{$86.62\%$}\\
\hline
\end{tabular}
\end{adjustbox}
\vspace{-5pt}
\end{table*}

\subsection{Attack Performance}

Tables~\ref{tab:transfer_attack},~\ref{tab:pla_softplus} and ~\ref{tab:pca} present the results of our studies on three types of attacks TAs, PLAs and PCAs, respectively. From the results, we have the following observations.

\noindent\textbf{Transfer attacks.}
As shown in Table~\ref{tab:transfer_attack}, the accuracy, F1, precision, and recall values are consistently high for most attacks. Specifically, the lowest attack accuracy is $75.66\%$. These high attack rates support our claim that KGE models are vulnerable to MIA attacks. In addition, we introduce the overfit level to indicate the target model's characteristics for ease of analysis. The overfit level is defined as follows:
 \vspace{-10pt}
\begin{equation}\label{eq:overfit_level}
    \textit{overfit level} = \textit{train acc} - \textit{test acc}
    \vspace{-10pt}
\end{equation}
By comparing attack performance in Table~\ref{tab:transfer_attack} and Figure~\ref{fig:ori_performance}, we observe the attack performance relates to overfit level and size of the dataset. Specifically, the average attack accuracy on five dataset are $93.02\%$, $81.91\%$, $85.42\%$, $97.49\%$, and $95.22\%$.  The FB15K237 and NELL-995 have more relations and triples, which is more difficult to overfit. Thus the adversary achieves lower attack accuracy. DDB14 has comparably more triples than FinKG, but DistMult achieves the lowest overfit level on the FinKG dataset as shown in Figure~\ref{fig:ori_performance}, which results in the lower average attack accuracy on the FinKG dataset.
On the other hand, by comparing the performance across models within the same dataset, we also find that the attack accuracy is highly correlated with overfit levels. For example, the attacks against TransE achieve the lowest accuracy on WN18RR, and TransE has the lowest overfit level on this dataset comparing other models. We notice similar signals for attacks against DistMult on FB15K237, TransE on NELL-995, and DistMult on FinKG, which satisfy our intuition that overfitting makes it distinguishable for training datasets and unseen datasets. The attacks against TransE and TransH on DDB14 are similar, and the overfit levels are comparable.

\begin{table*}[t]
\caption{Experimental results on PCAs.}\label{tab:pca}
\begin{adjustbox}{width=\textwidth,center}
    \centering
\begin{tabular}{|c|cccc|cccc|cccc|cccc|cccc|cccc}
\hline
{}& \multicolumn{4}{|c|}{WN18RR} &
\multicolumn{4}{c|}{FB15K237} &
\multicolumn{4}{c|}{NELL-995} &
\multicolumn{4}{c|}{DDB14}&
\multicolumn{4}{c|}{FinKG}\\
\hline
Model & Acc & F1 &Precision & Recall& Acc & F1 &Precision & Recall& Acc & F1 &Precision & Recall& Acc & F1 &Precision & Recall&Acc & F1 &Precision & Recall\\ 
\hline

TransE   &    69.81 &   76.81 &          62.35 &           100 &      57.60 &     70.22 &            54.11 &             100 &      62.78 &     71.43&            57.96&         93.07 &   64.15 &  73.61&         58.24&        99.99 &        62.68 &       72.8 &              57.26&               100 \\
TransH   &    77.62 &   81.71 &          69.08 &           100 &      60.26&     71.56 &            55.71 &             100 &      65.70 &     74.02 &            59.56 &         97.74 &   61.79 &  72.35 &         56.68 &          100 &        64.35 &       73.71&              58.37&               100 \\
ComplEx  &    87.70 &   89.04 &          80.25 &           100 &      59.80 &     71.33 &            55.43 &             100 &      69.72 &     76.75 &            62.28 &             100 &   69.20 &  76.45&         61.88&          100 &        61.49&       72.19 &              56.49 &               100 \\
DistMult &    87.77 &    89.10 &          80.35 &           100 &       56.83 &     69.84 &            53.66 &             100 &      62.61 &     72.78&            57.21&             100 &   67.47 &  75.45 &         60.58 &          100 &        56.36 &       69.61&              53.39 &               100 \\
\hline
{Average Acuracy}& \multicolumn{4}{|c|}{$80.62\%$} &
\multicolumn{4}{c|}{$58.62\%$} &
\multicolumn{4}{c|}{ $65.20\%$} &
\multicolumn{4}{c|}{$64.90\%$}&
\multicolumn{4}{c|}{$61.22\%$}\\
\hline
\end{tabular}
\end{adjustbox}
\vspace{-5pt}
\end{table*}

\begin{figure}[!ht]
    \centering
    \subfloat[ComplEx]{\includegraphics[width=2.5in]{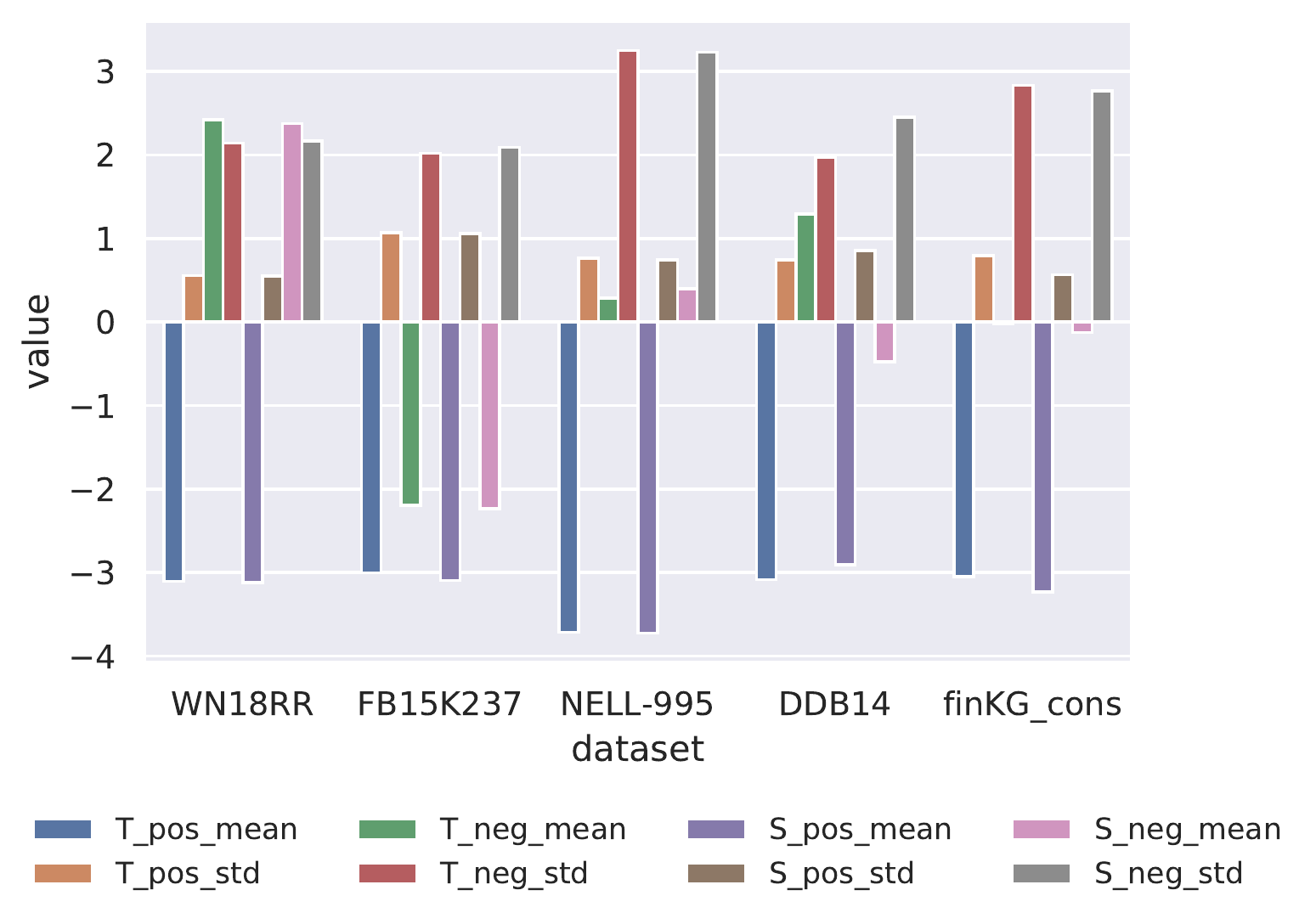}}
    
    \centering
    \subfloat[DDB14]{\includegraphics[width=2.5in]{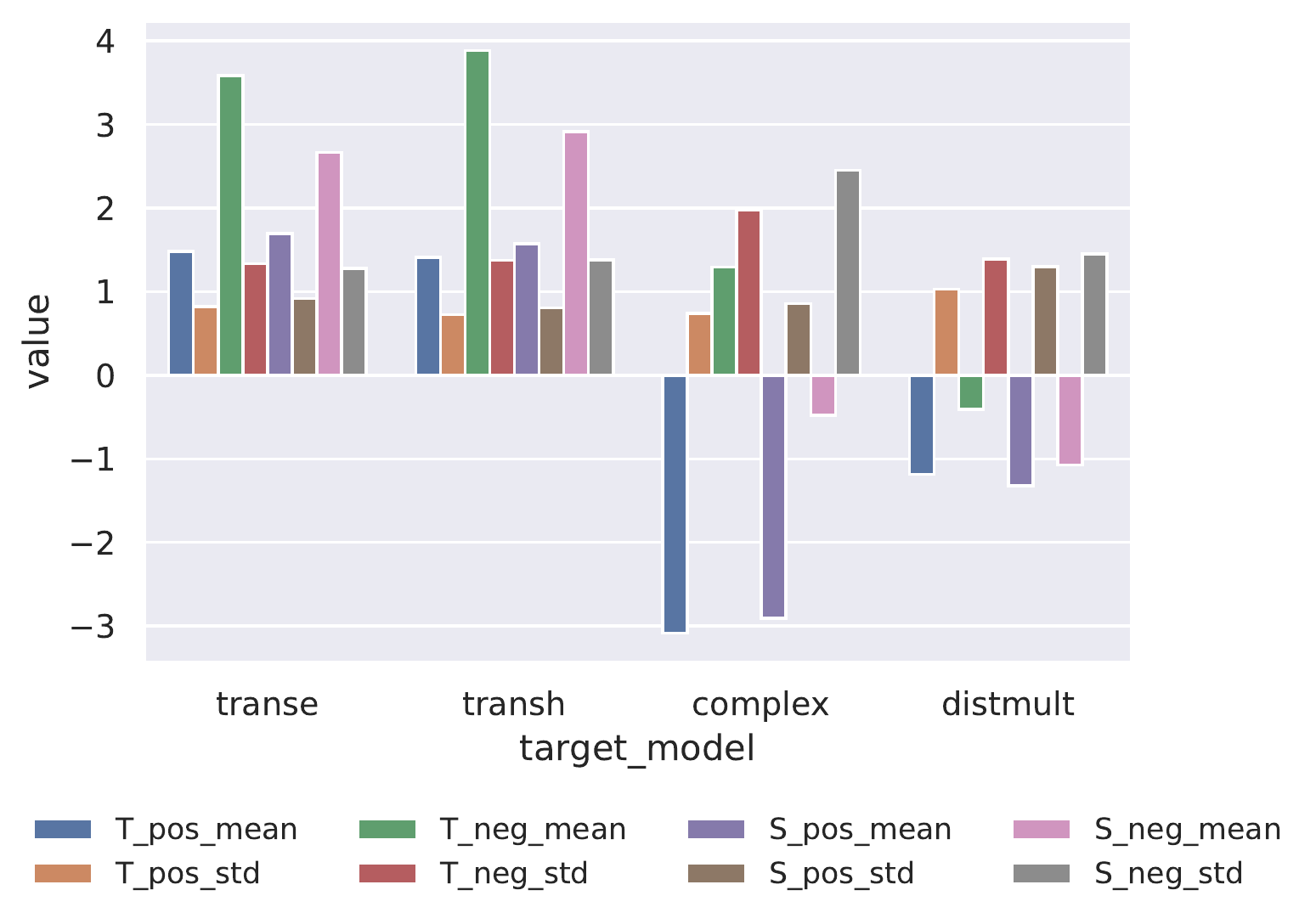}}
    \caption{The plausibility score analysis for ComplEx and DDB14. T, S, pos and neg indicate target, shadow, member samples and non-member samples respectively.}
    \label{fig:ori_score_ddb}
\vspace{-12pt}
\end{figure}

\noindent\textbf{Prediction loss-based attacks.}
In this scenario, we assume that the adversary only has access to the plausibility score from the target model. The shadow dataset is not available. Since there are two loss functions, we perform both attacks using these two loss functions as the metric. We report the attack performance in Table~\ref{tab:pla_softplus}, using logistic loss as the metric. The average attack accuracies on five datasets are $91.35\%$, $76.76\%$, $80.62\%$, $94.18\%$, and $92.50\%$, which only slightly degrade compared to TAs. These results demonstrate that the adversary can achieve successful membership inference attacks even without the help of shadow datasets. 
In Figure~\ref{tab:pla_margin}, we report the attack performance using margin loss as the metric.
By comparing the attack accuracy of four models on these five datasets across margin loss and logistic loss, we observe that the attack performance using logistic loss is slightly better than that using margin loss as the metric. This observation indicates that logistic loss is more sensitive to membership inference attacks. We observe that the attack performance improves largely when changing the loss metric from margin to logistic loss, especially for TransE and TransH. For example, the PLAs against TransE on DDB14 with margin loss metric is $81.32\%$, while the performance improves to $94.76\%$ when we change the metric as the logistic loss. 
We observe in Figure~\ref{fig:ori_score_ddb}(b) that the score function of TransE and TransH will give both true triples and corrupted triples a positive value, and the non-linearity of logistic based metric makes scores more distinguishable between trained triples and test triples.

\noindent\textbf{Prediction correctness-based attacks.}
In PCAs, the adversary can only obtain the predicted hard labels from the target model. From Table~\ref{tab:pca}, we can see that the attack accuracy of PCAs degrades compared to PLAs. Specifically, the average PCA performances on each dataset are $80.62\%$, $58.62\%$, $65.20\%$, $64.90\%$, and $61.22\%$ respectively. PCAs performance generally degrades compared to TAs and PLAs. The main reason is that hard labels provide the least information for the adversary. By comparing the PCAs performance and the target model performance, we observe that the PCAs performance is highly correlated to recall of target models. To be detailed, as shown in Tabel~\ref{tab:pca}, the PCAs against Distmult on WN18RR, TransH on FB15K237, ComplEx on NELL-995, ComplEx on DDB14, and TransH on FinKG achieve the highest attack accuracy, while the corresponding target model on such datasets achieve the lowest recalls. Since low recall indicates that only a small fraction of true triples are predicted correctly, the adversary predicts a large fraction of true triples in the test dataset as negative samples, which results in high attack performance.

Furthermore, Table~\ref{tab:pca} shows that the attack model has high recall and low precision in general. This observation indicates that the attacks could correctly predict most training samples but wrongly label some test samples as training set members, further demonstrating privacy issues.

\subsection{Ablation Study}
This section conducts ablation studies to understand better the effects of different components and assumptions on attack performance.
Previously, we assume that the shadow dataset should have similar data distribution as the target training dataset. However, this assumption is rigid to guarantee in some cases. For example, the financial companies train their KGE models with a private KG and will not publish any part of this KG. Therefore, it is hard for the adversary to obtain such a dataset from these institutes. Thus, we perform attacks by relaxing the above assumption. For a fair comparison, we conduct attacks against models that are trained 100 times, and embedding dim is 200. 
\begin{figure}
    \centering
    \includegraphics[width=2.5in]{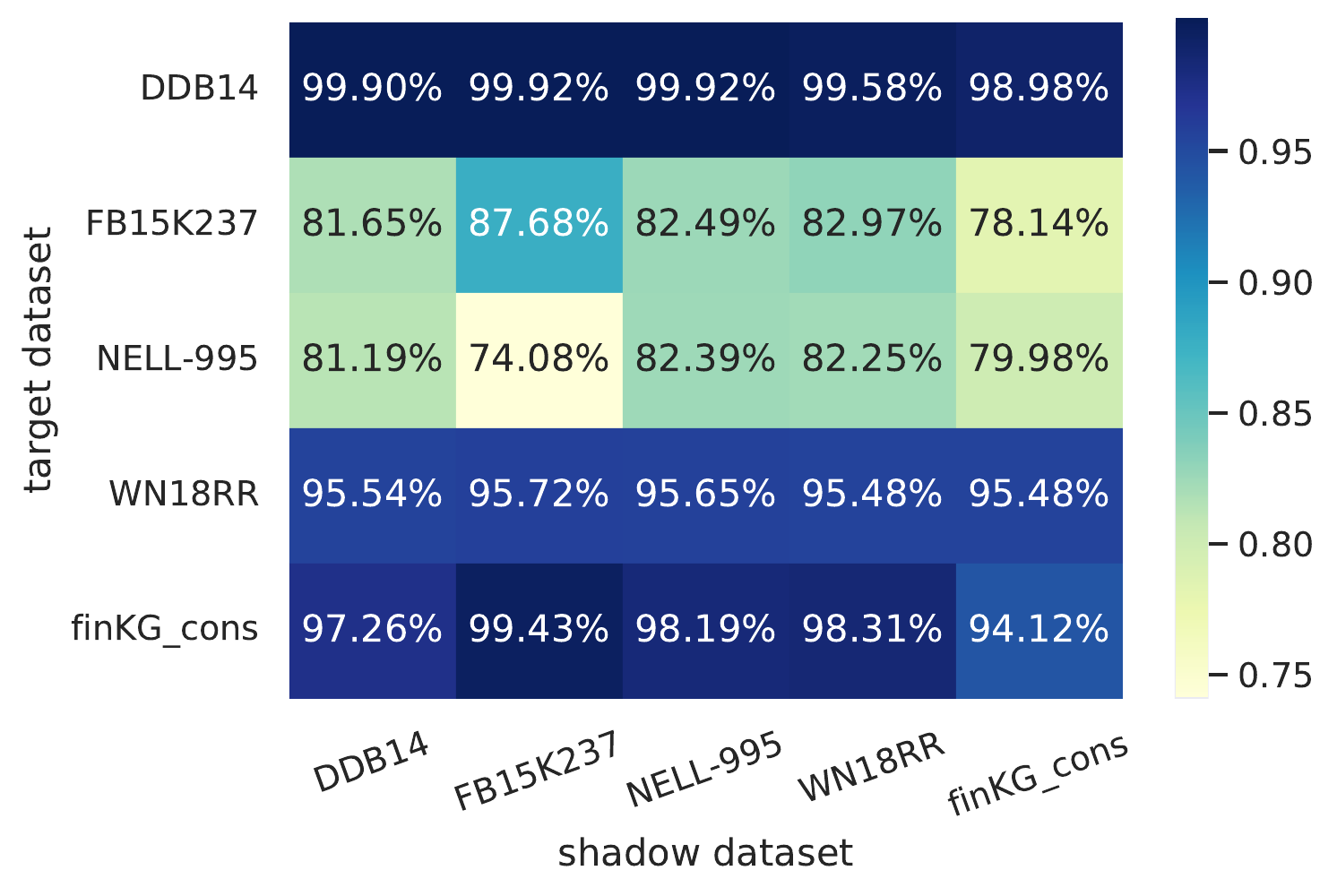}
    \caption{Transfer attacks across dataset for ComplEx}
    \label{fig:ad}
\vspace{-15pt}
\end{figure}

\noindent\textbf{Attacks across different datasets.}
In Figure~\ref{fig:ad}, we use ComplEx as our model. We conduct TAs with different target datasets and shadow datasets. The attacks are still successful since the attack accuracy is above $74.08\%$ in all cases. These results indicate that the adversary can successfully retrieve membership information of the private target dataset even without the assistance of a similar dataset. From Figure~\ref{fig:ad}, we observe that the attack performance is strongly contingent on the target dataset, while only having a small correlation to the choice of shadow dataset. Specifically, the FB15K237 and NELL-995 are the two largest datasets with the lowest attack accuracy. We observe (shown in Figure~\ref{fig:ori_score_ddb}(a))that the target model assigns a small negative value for training samples regardless of the datasets. Thus, the shadow dataset is only used to help the adversary find the target model's decision boundary. It is worth noting that attack performance against ComplEx on FinKG is more successful using FB15K237, NELL-995, and WN18RR as shadow datasets than attacks using FinKG as a shadow dataset. The possible reason is that a large shadow dataset gives the adversary more details about ComplEx's decision boundary.

\begin{figure}
    \centering
    \includegraphics[width=2.3in]{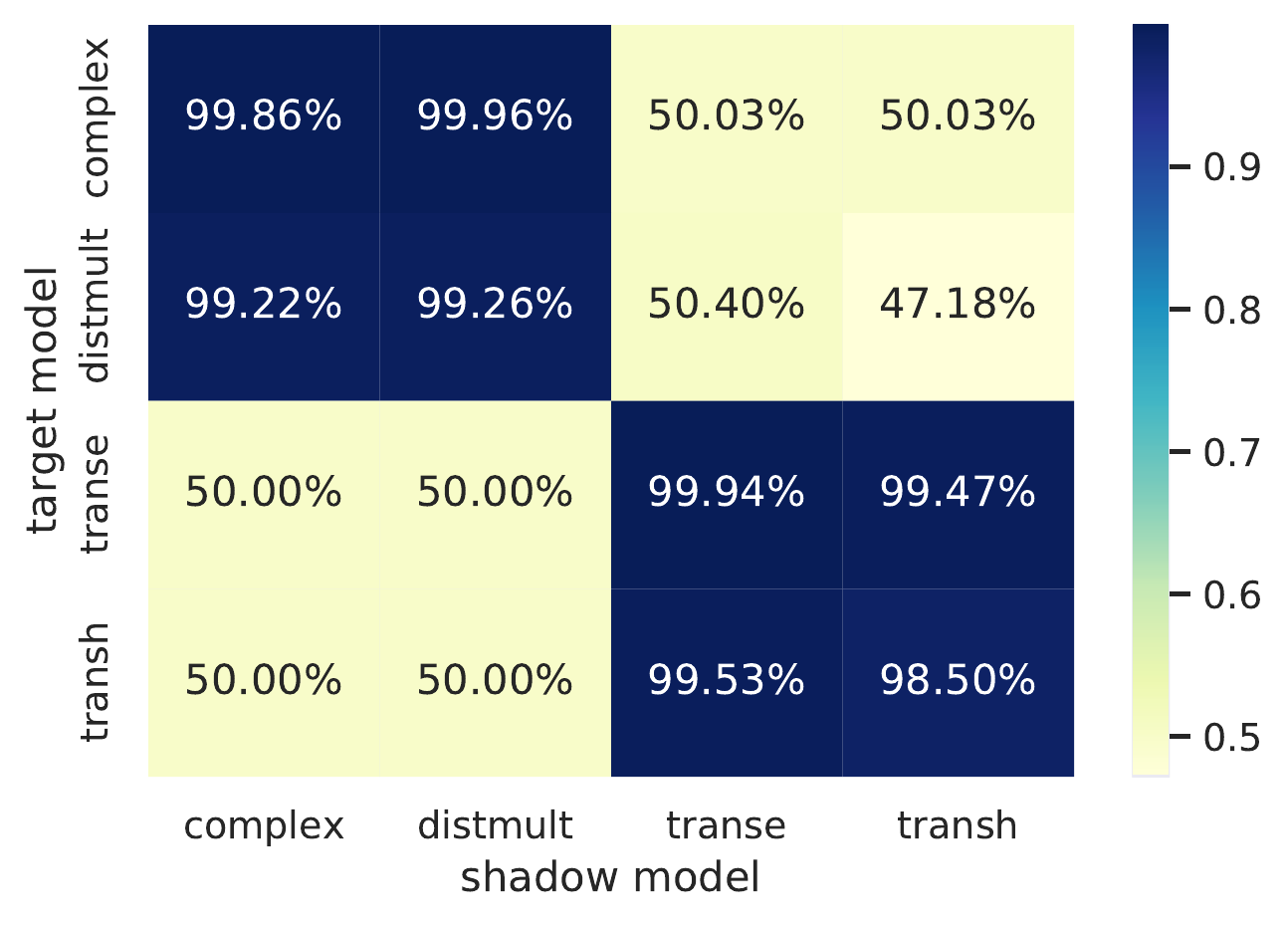}
    \caption{TAs across models on DDB14}
    \label{fig:am}
\vspace{-15pt}
\end{figure}

\noindent\textbf{Attacks across different models.}
We further perform transfer attacks by relaxing the assumption that the shadow model should have the same architecture as the target model. In Figure~\ref{fig:am}, we report the transfer attack accuracy against four target models on DDB14 applying different shadow model architecture. 
The results show that the attack obtains reasonable accuracy when the shadow model and target model are trained using the same loss function. For example, the attack accuracy against TransE is $99.47\%$ when the shadow model is TransH. On the contrary, the attack against TransH fails when the shadow model is DistMult. The attack accuracy is $50\%$, comparable with a random guess.
This phenomenon results from the different scoring behaviors between these two models. We visualize the analysis of scores for other models on DDB14 in Figure \ref{fig:am}. We observe that margin-based models (TransE and TransH) assign a positive score to training samples, while logistic-based models (ComplEx and DistMult) assign negative scores to training samples. This different scoring behavior makes it hard for the adversary to retrieve a reasonable decision boundary of the target model. The logistic-based shadow model cannot mimic behaviors of the margin-based target model and vice versa.

\section{Related Work}

In this section, we review some recent works related to our inference attacks in three perspectives, including membership inference attacks, privacy attacks on graphs, and knowledge graph methods.

\noindent\textbf{Membership Inference Attacks} \cite{shokri2017membership,hayes2017logan,nasr2018machine,salem2018ml, chen2018differentially,yeom2018privacy,nasr2019comprehensive,carlini2019secret,song2019robust,li2020label} support an adversary in inferring whether a data sample serves as a training sample for the target model. \citet{shokri2017membership} propose the first MIAs against machine learning models and examine their relationship with model overfitting.
\citet{salem2018ml} propose model and data-independent MIAs via relaxing the assumptions made by \citet{shokri2017membership}. The reveal of privacy leakage on euclidean datasets, e.g., images, motivates the investigation on non-euclidean datasets, such as graphs.\\ 
\noindent\textbf{Privacy Attacks on Graphs.}
~\citet{ellers2019privacy} measure the changes of node embeddings after removing the target node to infer the link information of the target node.~\citet{duddu2020quantifying, olatunji2021membership} conduct black-box MIAs to infer whether a node belongs to a training graph by analyzing confidence scores. The node embeddings from confidence scores can also be retrieved by solving a linear system if prediction scores are well-scaled. However, this assumption is not realistic for KGE attacks since prediction scores of KGE methods are scale-free and entity and relation embeddings are unavailable.~\citet{duddu2020quantifying} train a graph autoencoder to reconstruct graphs with target node embedding as input.~\citet{he2021node} perform MIAs by analyzing confidence scores queried from the target model, assuming access to the knowledge of node neighbors. However, the target model knowledge access assumptions of these prior works are too loose and we only obtain the symbolic index of a candidate triple in real-world setting.

\noindent\textbf{Knowledge Graph Embedding} methods have been extensively studied in recent years. Most previous work~\cite{bordes2011learning,bordes2013transE,wang2014knowledge,yang2014embedding,lin2015learning,nguyen2016stranse,trouillon2016complex,nickel2016holographic,dettmers2018convolutional,sun2019rotate} follow the general methodology of learning continuous vectors for entities and relations by maximizing the salience of a collection of factual triplets $(h, r, t)$. For example, TransE~\cite{bordes2013transE}, TransH~\cite{wang2014knowledge} and TransR~\cite{lin2015learning} all assume each relation $r$ as a transformation from a head entity $h$ to a tail entity $t$, while ComplEx~\cite{trouillon2016complex} and RotatE~\cite{sun2019rotate} further extend them by introducing complex embeddings to better model asymmetric relations. During learning, all previous studies do not consider sensitive information from the KGs, thus potentially leading to privacy leakage.

\section{Conclusion}

In this paper, we propose three different types of membership inference attacks (MIAs) on knowledge graphs according to the scope of the adversary's knowledge.
The adversary can explore private information with limited knowledge, which reveals the high risk of privacy leakage for the KGE models.
Our empirical results demonstrate that the performance of MIA on knowledge graphs is very significant, which alerts the privacy leakages from the KGE models. We hope this work can considerably accelerate the privacy-preserving KGE methods against privacy attacks.

\bibliography{anthology,custom}
\bibliographystyle{acl_natbib}

\newpage
\appendix

\end{document}